\documentclass[conference,a4paper]{IEEEtran}
\IEEEoverridecommandlockouts

\usepackage[hidelinks]{hyperref}
\usepackage[cmex10]{amsmath}
\usepackage{amssymb,amsfonts}
\usepackage{dblfloatfix}

\usepackage[ruled,vlined]{algorithm2e}
\usepackage{graphicx}
\usepackage{graphics}
\usepackage{subcaption} 
\usepackage{multirow}
\graphicspath{{Figures/PDF/}{Figures/PNG/}}

\usepackage{booktabs}
\usepackage{siunitx}
\usepackage[numbers,compress]{natbib}
\usepackage{texnames}
\usepackage{bm,bbm}
\usepackage{orcidlink}
\usepackage{epsfig}
\usepackage{adjustbox}
\usepackage{setspace}

\begin{document}

\title{Beyond Task-Driven Features for Object Detection}

\author{
    \IEEEauthorblockN{Meilun Zhou\textsuperscript{1}\orcidlink{0000-0002-5891-3203}}
    \IEEEauthorblockA{\textsuperscript{1}Department of Electrical and Computer Engineering\\
    \textit{University of Florida}\\
    Gainesville, FL, USA, 32611\\
    zhou.m@ufl.edu}
    \and
    \IEEEauthorblockN{Alina Zare\textsuperscript{1,2}\orcidlink{0000-0002-4847-7604}}
    \IEEEauthorblockA{\textsuperscript{1}Department of Electrical and Computer Engineering\\
    \textsuperscript{2}Artificial Intelligence and Informatics Research Institute\\
    \textit{University of Florida}\\
    Gainesville, FL, USA, 32611\\
    azare@ufl.edu}
}


\maketitle
\begin{abstract}
Task-driven features learned by modern object detectors optimize end task loss yet often capture shortcut correlations that fail to reflect underlying annotation structure. Such representations limit transfer, interpretability, and robustness when task definitions change or supervision becomes sparse. This paper introduces an annotation-guided feature augmentation framework that injects embeddings into an object detection backbone. The method constructs dense spatial feature grids from annotation-guided latent spaces and fuses them with feature pyramid representations to influence region proposal and detection heads. Experiments across wildlife and remote sensing datasets evaluate classification, localization, and data efficiency under multiple supervision regimes. Results show consistent improvements in object focus, reduced background sensitivity, and stronger generalization to unseen or weakly supervised tasks. The findings demonstrate that aligning features with annotation geometry yields more meaningful representations than purely task optimized features.
\end{abstract}


\section{Introduction}
Two-stage object detectors such as Faster R-CNN \cite{ren2016faster} rely on a shared convolutional backbone and a Feature Pyramid Network (FPN) \cite{lin2017feature} whose outputs provide the primary input to both the Region Proposal Network (RPN) and the region-based prediction heads \cite{he2017mask}. While this shared representation supports multiple objectives, optimization typically occurs through detection-specific losses such as cross-entropy for classification and smooth $L_{1}$ for bounding box regression. Previous studies suggest that these task-driven gradients prioritize discriminative boundaries over the underlying structural properties of the objects \cite{locatello2020object}. Consequently, the resulting feature hierarchy reflects narrow task-specific optimization rather than an object-centric structure grounded in how objects relate across scales, which is a limitation that current research in small-object detection identifies as a primary cause of feature degradation in complex scenes \cite{aldubaikhi2025advancements}.

The backbone and FPN together form the representational bottleneck of the Faster R-CNN architecture. Because the RPN and detection heads draw from the same feature pool, any inductive bias introduced at this stage influences both proposal generation and final prediction \cite{lin2017feature}. Modifying downstream heads can improve local reasoning, but cannot alter the shared representation that governs initial spatial localization \cite{bello2024reprot}. Recent advancements in structure-controllable models demonstrate that augmenting FPN features with object-aligned structure such as spatial layouts or instance masks provides a more principled mechanism for improving both proposal quality and final predictions without overhauling the entire detection pipeline \cite{fang2024data,zhu2025recon}.

\begin{figure}[htpb]
    \centering
    \includegraphics[width=0.7\linewidth]{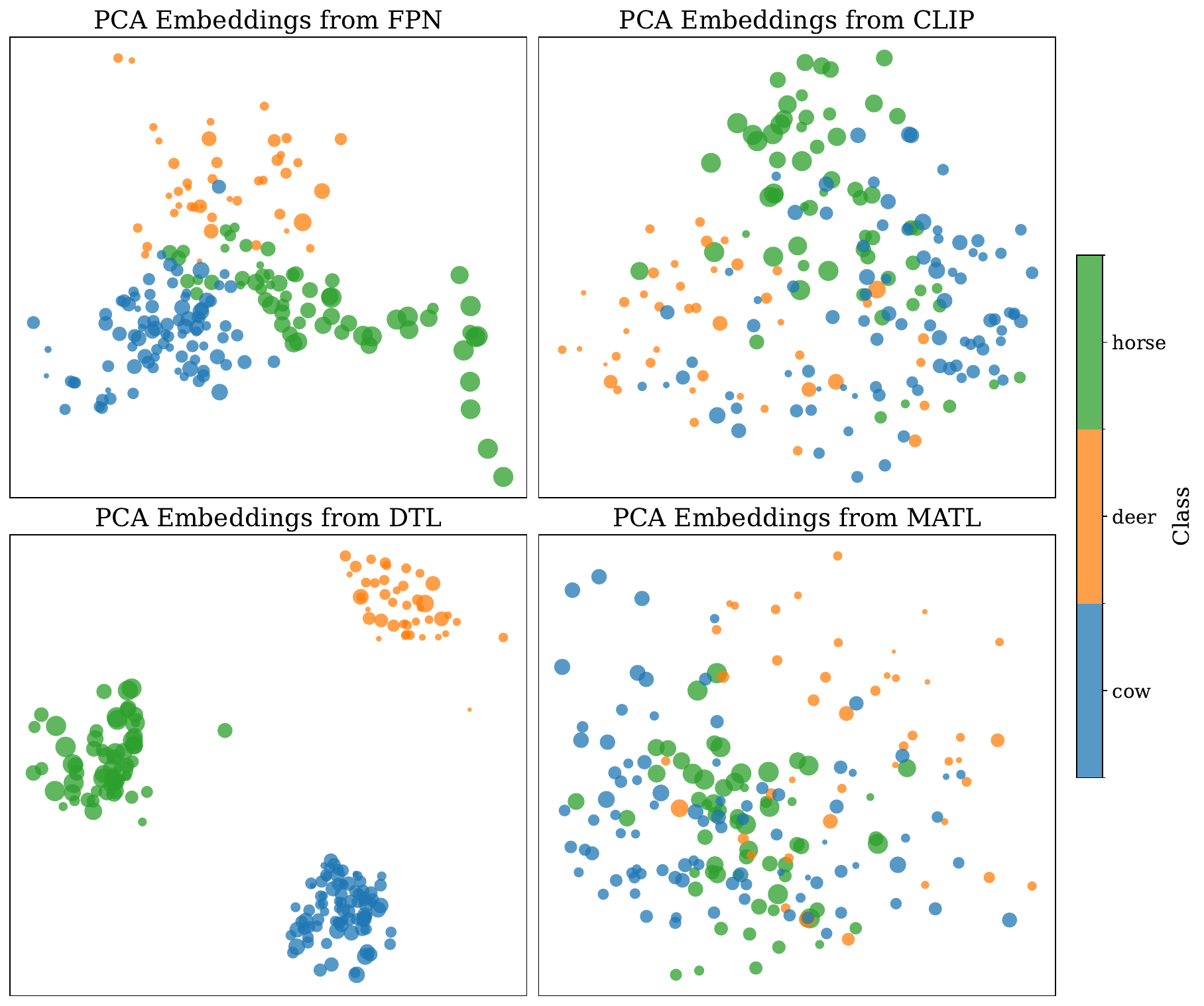}
    \caption{
        Two-dimensional PCA projections of embeddings extracted from the Faster R-CNN backbone, CLIP, DTL, and MATL. The MATL embedding separates class labels while preserving intra-class structure.
    }
    \label{fig:embedding_figure}
\end{figure}

Latent spaces for representation learning should aim to capture how real objects relate to one another rather than reflecting geometry induced solely by loss-driven gradients. Multi-Annotation Triplet Loss (MATL) \cite{zhou2025multi} provides an initial framework for constructing such spaces by incorporating additional annotations such as bounding box area and symmetric squareness alongside class labels. Instead of encoding similarity as a byproduct of class separation, MATL explicitly organizes embeddings according to continuous relationships derived from multiple attributes. An example of this can be seen in Fig. \ref{fig:embedding_figure}. This design preserves structured variation across class identity and object geometry, which results in a shared and discriminative manifold where objects of similar spatial extent cluster together regardless of semantic category \cite{dutt2022shared}. 

This work introduces a backbone-level augmentation strategy that injects dense multi-annotation latent features into Feature Pyramid Network representations. A frozen and task-independent latent encoder integrates into a two-stage detector without modifying the region proposal or region-based prediction heads, serving as a first-pass validation of the approach. Although the empirical evaluation focuses on Faster R-CNN, the proposed concept is architecture-agnostic and applies naturally to state-of-the-art one-stage and transformer-based detectors where backbone representations form the primary bottleneck for object understanding. The observed improvements in proposal quality and final detection performance support the broader premise that annotation-guided latent priors can act as reusable representational components across modern detection frameworks.



\section{Proposed Methodology}
This section presents a unified methodology for constructing and deploying an object-aligned latent space within object detection pipelines. The approach begins by learning a multi-annotation latent representation from object-centric image patches using a triplet-based objective that jointly encodes semantic and geometric structure. That learned representation is then extended to full detection scenes through a sliding-window formulation that produces dense spatial grids of latent features. Finally, these grids are integrated into a standard object detector through channel-wise fusion, allowing proposal generation and region-based prediction to exploit latent structure learned independently of the detection objective. Together, these components define a two-stage framework that enables the learned latent geometry to influence both region proposals and final detection decisions.

\subsection{Learning the Multi-Annotation Latent Space}
The multi-annotation triplet loss (MATL) \cite{zhou2025multi} is currently designed to operate on images that contain at most a single object or no object at all. A pre-trained CLIP \cite{radford2021learning} foundation model is used as the backbone for representation learning. Each image is first passed through the CLIP encoder to obtain a fixed-length embedding, and these embeddings serve as the inputs to the multi-annotation triplet loss during training. Each image is paired with a continuous annotation vector that encodes the properties and spatial attributes of one object or explicitly represents background structure. The MATL loss shapes the latent space by enforcing distances between embeddings according to these annotation vectors. Training is performed on single-object and background-only images so that each embedding corresponds to a single object in the latent space.

\subsection{Dense Latent Feature Grid Construction}

Let $\mathbf{I}\in\mathbb{R}^{3\times H\times W}$ denote an input image for an object detection task. The embedding network $f(\cdot)$ trained using multi-annotation triplet loss is repurposed as a fixed patch encoder and applied to $\mathbf{I}$ through a sliding-window operator with fixed window size and stride. Each window defines a localized image region, and the encoder produces a latent embedding vector that reflects object-aligned semantic and geometric structure learned during representation training.

Evaluating the encoder over all window locations yields a dense latent feature grid $\mathbf{G} \in \mathbb{R}^{D \times H_g \times W_g}$, where $D$ denotes the embedding dimension and $(H_g, W_g)$ denotes the spatial resolution induced by the sliding-window configuration. Each grid location corresponds to a specific image region and encodes localized latent structure that is not explicitly optimized by the detection objective.

The spatial resolution of $\mathbf{G}$ generally differs from the resolutions of the Feature Pyramid Network (FPN) feature maps. To enable integration, the grid is spatially aligned to each pyramid level. For a given FPN level $\ell$ with spatial resolution $(H_\ell, W_\ell)$, a resampling operator
\[
\mathcal{R}_\ell:\mathbb{R}^{D \times H_g \times W_g} \rightarrow \mathbb{R}^{D \times H_\ell \times W_\ell}
\]
maps the latent grid to the target resolution,
\begin{equation}
\widetilde{\mathbf{G}}_\ell = \mathcal{R}_\ell(\mathbf{G}).
\end{equation}
In practice, $\mathcal{R}_\ell(\cdot)$ is implemented using bilinear interpolation. This operation preserves spatial correspondence between latent embeddings and backbone activations at each pyramid level.

\subsection{Backbone Feature Pyramid Augmentation}

Let $\mathbf{P}_\ell \in \mathbb{R}^{C \times H_\ell \times W_\ell}$ denote the feature map produced by the Feature Pyramid Network (FPN) at pyramid level $\ell$, where $C=256$ for the ResNet50--FPN backbone and $(H_\ell, W_\ell)$ denote the spatial resolution at that level. Let $\widetilde{\mathbf{G}}_\ell \in \mathbb{R}^{D \times H_\ell \times W_\ell}$ denote the latent feature grid obtained by spatially aligning the multi-annotation embedding grid to the resolution of level $\ell$, where $D$ is the latent grid dimensionality.

To enable fusion with the backbone features, the latent grid is first projected into the FPN channel space using a learnable $1\times1$ convolution,
\begin{equation}
\mathbf{G}_\ell = \psi_\ell(\widetilde{\mathbf{G}}_\ell),
\qquad
\mathbf{G}_\ell \in \mathbb{R}^{C \times H_\ell \times W_\ell},
\end{equation}
where $\psi_\ell : \mathbb{R}^{D \times H_\ell \times W_\ell} \rightarrow \mathbb{R}^{C \times H_\ell \times W_\ell}$ is applied independently at each pyramid level.

We consider three strategies for fusing the backbone features $\mathbf{P}_\ell$ with the projected latent features $\mathbf{G}_\ell$.

\paragraph{Additive fusion}
Additive fusion performs element-wise summation of the two feature maps,
\begin{equation}
\mathbf{P}_\ell^{\mathrm{aug}} = \mathbf{P}_\ell + \mathbf{G}_\ell,
\end{equation}
which injects latent structure directly into the backbone representation while preserving the original channel dimensionality.

\paragraph{FiLM fusion}
Feature-wise Linear Modulation (FiLM) \cite{turkoglu2022film} conditions the backbone features using channel-wise affine transformations predicted from the latent grid,
\begin{equation}
\mathbf{P}_\ell^{\mathrm{aug}} =
\boldsymbol{\gamma}_\ell \odot \mathbf{P}_\ell + \boldsymbol{\beta}_\ell,
\end{equation}
where $\odot$ denotes element-wise multiplication, and $(\boldsymbol{\gamma}_\ell, \boldsymbol{\beta}_\ell) \in \mathbb{R}^{C \times H_\ell \times W_\ell}$ are modulation parameters obtained from $\mathbf{G}_\ell$ via learnable $1\times1$ convolutions.

\paragraph{Spatial mask fusion}
Spatial mask fusion uses the latent grid to produce a spatially varying attention mask that modulates the backbone features,
\begin{equation}
\mathbf{M}_\ell = \sigma\!\left(h_\ell(\mathbf{G}_\ell)\right),
\qquad
\mathbf{M}_\ell \in \mathbb{R}^{1 \times H_\ell \times W_\ell},
\end{equation}
where $h_\ell$ is a learnable $1\times1$ convolution and $\sigma(\cdot)$ denotes the sigmoid function. The augmented features are then given by
\begin{equation}
\mathbf{P}_\ell^{\mathrm{aug}} = \mathbf{M}_\ell \odot \mathbf{P}_\ell.
\end{equation}

For pyramid levels that are not selected for augmentation, the original features are passed through unchanged. The augmented pyramid features $\{\mathbf{P}_\ell^{\mathrm{aug}}\}$ replace the original FPN outputs throughout the Faster R-CNN pipeline. The RPN operates directly on $\mathbf{P}_\ell^{\mathrm{aug}}$ to generate object proposals, and RoIAlign \cite{bai2020optimized} samples from the same augmented feature maps during region-wise classification and bounding box regression. 



\section{Experimental Results}
The dataset for our experiments are derived from the Animal Wildlife Image Repository (AWIR) dataset used in \cite{krishnan2023fusion}. The original data preparation pipeline from the MATL paper \cite{zhou2025multi} including image tiling, target isolation, annotation processing, and feature extraction is adopted here to ensure consistency across all experiments. MATL networks are trained for $1000$ epochs and Faster R-CNN networks are trained for $200$ epochs. All networks are repeated $5$ times across different training and validation splits to check for consistency. 

All models were trained with Tensorflow on an NVIDIA B200 GPU with $12$ CPUs and $60$ GB of memory allocated. Code can be found at \url{https://github.com/GatorSense/Annotation-Driven-Detection}.

\begin{figure}[htpb]
    \centering
    \includegraphics[width=0.7\linewidth]{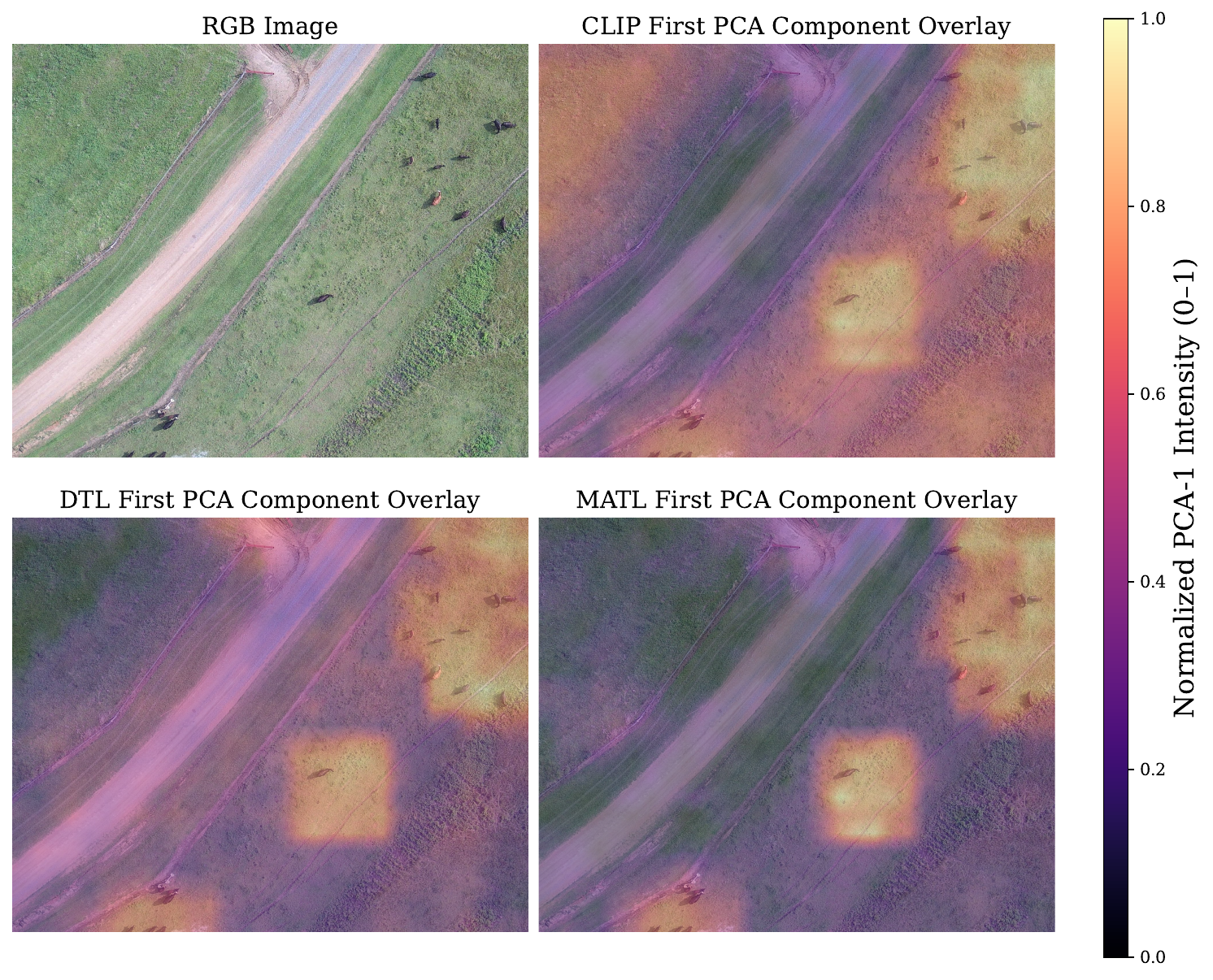}
    \caption{
        Brighter regions indicate higher values of the first principal component computed from the dense latent feature grids produced by sliding-window embedding and backbone-aligned projection. MATL concentrates activation on object-relevant regions more consistently than CLIP or DTL which both exhibit stronger responses to background areas.
    }
    \label{fig:pca_figure}
\end{figure}


\begin{figure*}[htpb]
    \centering
    \includegraphics[width=0.9\linewidth]{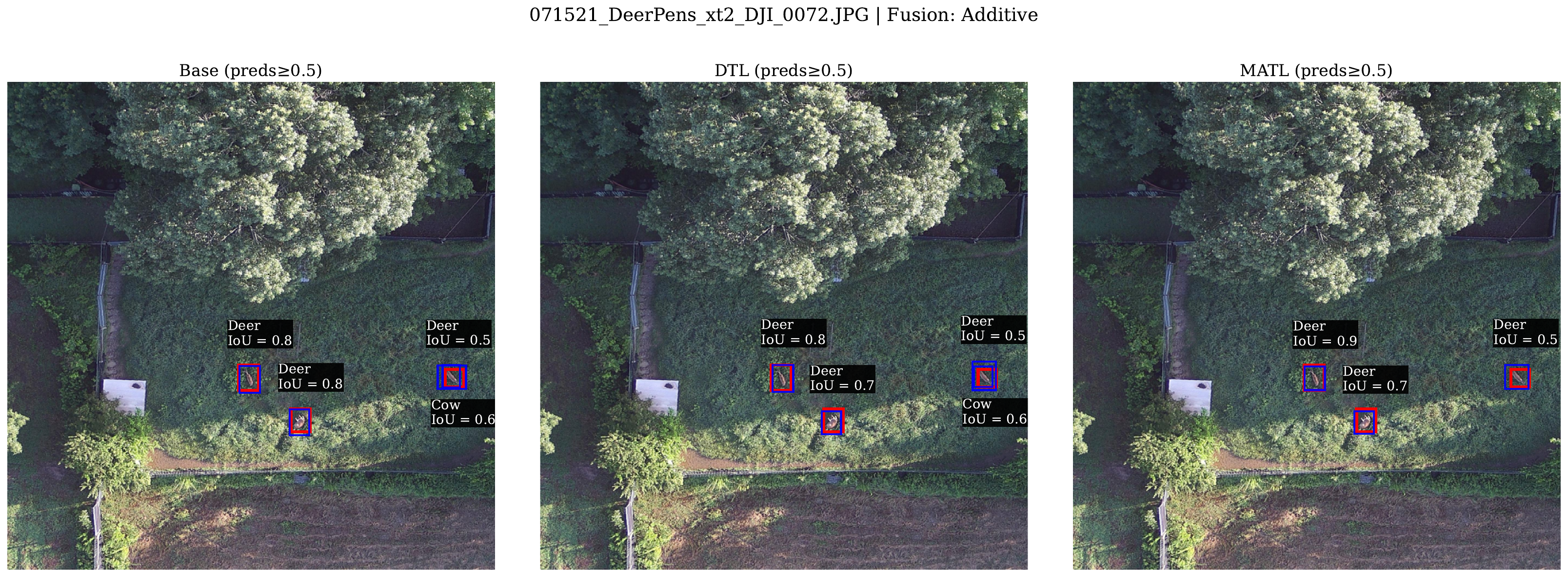}
    \caption{
        In this scene, MATL correctly localizes and classifies the three deer, while both the baseline and DTL models additionally produce a false positive cow prediction. Faster R-CNN augmented with MATL also produces boxes with higher IoU with the ground truth. The predictions all have a confidence threshold of 0.5 or higher.
    }
    \label{fig:detections}
\end{figure*}

For object detection experiments, the dataset is split into $60$\% training, $20$\% validation, and $20$\% test sets using random sampling with class stratification. Object detection models are implemented using Faster R-CNN with a ResNet-50 backbone \cite{ren2016faster} and a Feature Pyramid Network (FPN). The FPN produces multi-scale feature maps with 256 channels at each pyramid level. Latent grid features derived from the MATL embeddings are spatially aligned and fused into the FPN at levels ${P_2, P_3, P_4, P_5}$, which correspond to progressively coarser spatial resolutions.

The MATL model is trained using the procedure introduced in \cite{zhou2025multi}, with one modification to the original protocol. Image patches are generated from only the training and validation split used for object detection. Background-only patches are included as an explicit class. For these samples, the bounding box spans the full patch extent, which avoids ambiguous geometric supervision. During inference over full detection scenes, a sliding-window procedure extracts $300\times300$ pixel image regions with a stride of $50$ pixels. All embeddings are projected to a latent dimension of $64$ and the first principal component of each setup is visualized in Fig. \ref{fig:pca_figure}.

This study examines how annotation-guided latent features affect object detection in a fixed Faster R-CNN framework by comparing a baseline model, a detector augmented with discrete triplet loss (DTL) features, and a detector augmented with multi-annotation triplet loss (MATL) features derived from continuous semantic and geometric supervision. Additive, FiLM-based modulation, and spatial masking fusion are evaluated to determine how latent spaces should be integrated into a task-driven backbone. All models share the same architecture, training protocol, and optimization settings, with backbone weights and detection losses held constant.

\begin{figure}[htpb]
    \centering
    \includegraphics[width=0.7\linewidth]{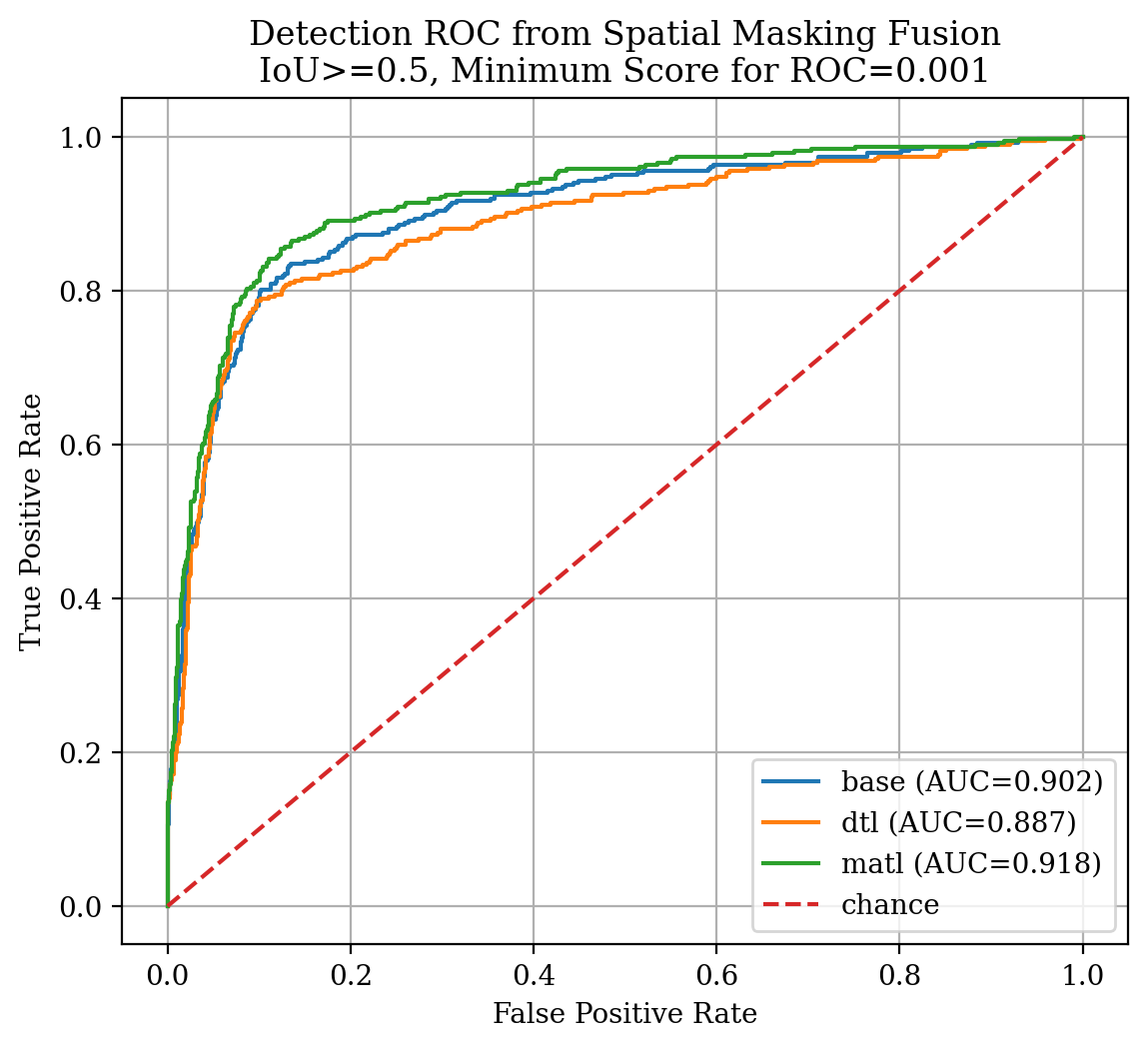}
    \caption{
        Receiver operating characteristic (ROC) curves comparing detection performance using base R-CNN, DTL-guided features, and MATL-guided features against a chance baseline. AUC scores are provided in the legend.
    }
    \label{fig:roc_curve}
\end{figure}

\newcommand{\pmstd}[1]{\mathbin{\vcenter{\hbox{$\scriptstyle \pm #1$}}}}

\begin{table}[!htb]
\caption{Detection performance on the AWIR dataset for different grid fusion and latent space techniques. Metrics report overall $\mathrm{mAP}_{50}$, precision (P), and recall (R) for Base, DTL, and MATL models. The best performing models for each metric are highlighted in bold.}
\centering
\setlength{\tabcolsep}{4pt}
\begin{tabular}{|c|c|c|c|c|}
\hline
\textbf{Method} & $\mathbf{mAP}_{50}$ & \textbf{P} & \textbf{R} & \textbf{F1}\\
\hline\hline
Base & $0.62\pmstd{0.03}$ & $0.63\pmstd{0.03}$ & $0.76\pmstd{0.03}$ & $0.69\pmstd{0.03}$ \\
\hline\hline

\multicolumn{5}{|c|}{\textbf{Spatial Mask Fusion}} \\
\hline
DTL  & $0.63\pmstd{0.06}$ & $0.62\pmstd{0.04}$ & $0.78\pmstd{0.03}$ & $0.69\pmstd{0.03}$\\
MATL & $\mathbf{0.69\pmstd{0.05}}$ & $0.60\pmstd{0.03}$ & $\mathbf{0.83\pmstd{0.05}}$ & $0.70\pmstd{0.03}$\\
\hline\hline

\multicolumn{5}{|c|}{\textbf{FiLM Fusion}} \\
\hline
DTL  & $0.63\pmstd{0.06}$ & $0.62\pmstd{0.04}$ & $0.77\pmstd{0.04}$ & $0.69\pmstd{0.04}$\\
MATL & $0.66\pmstd{0.07}$ & $\mathbf{0.65\pmstd{0.04}}$ & $0.79\pmstd{0.05}$ & $\mathbf{0.71\pmstd{0.04}}$\\
\hline\hline

\multicolumn{5}{|c|}{\textbf{Additive Fusion}} \\
\hline
DTL  & $0.67\pmstd{0.06}$ & $0.64\pmstd{0.03}$ & $0.80\pmstd{0.04}$ & $\mathbf{0.71\pmstd{0.03}}$\\
MATL & $0.67\pmstd{0.05}$ & $0.63\pmstd{0.02}$ & $0.80\pmstd{0.04}$ & $0.70\pmstd{0.02}$\\
\hline
\end{tabular}
\label{table:awir_detection_fusion}
\end{table}

Table \ref{table:awir_detection_fusion} compares detection performance across latent space variants and feature fusion strategies. The strongest results occur when MATL features are fused using spatial masking. This configuration yields the best overall detection quality, indicating that spatial masking enables location-aware integration of MATL features into the backbone. This behavior aligns with preserving object-aligned structure while suppressing background responses. Improved recall suggests that more true objects are proposed and retained throughout the detection pipeline. Across all fusion strategies, MATL consistently matches or outperforms DTL in detection performance. Fig. \ref{fig:roc_curve} shows that MATL-guided features achieves a higher AUC than the base Faster R-CNN which indicates improved discrimination across detection thresholds. Qualitative examples of improved detections appear in Fig. \ref{fig:detections}. 


Overall, the results highlight that performance gains depend jointly on the latent space design and the fusion mechanism. Spatial masking provides the most effective interface for MATL features. This strategy enables the detector to leverage object-aligned latent structure at the backbone level while maintaining strong recall and overall detection accuracy.

\section{Discussion}
A latent space learned from annotation-guided supervision encodes relationships that reflect semantic and geometric similarity rather than task-specific loss gradients. Such structure remains stable across downstream tasks because the representation is not shaped by proposal heuristics or classification requirements. The improvements observed when integrating these features indicate that a meaningful latent space provides information that may be absent from task-driven representations alone. Because this latent structure is learned independently of any specific detection architecture, the same principle can be applied to modern one-stage and transformer-based detectors where backbone features serve as the primary interface between perception and prediction.

Within this framework, DTL features improve detection performance by organizing features that emphasize inter-class separation. MATL further advances performance because the higher quality latent space jointly separates class identity and box-level geometry. This structure supplies the detector with features that are informative for both classification and localization, which results in stronger proposal quality and more accurate box refinement while preserving semantic discrimination.

Our sliding-window strategy offers a practical way to apply an object-centric latent space to full detection scenes, but this design has clear limitations. Each window summarizes a fixed spatial region with a single embedding, which may become ambiguous when multiple objects reside in the same window. Additionally, representation quality depends on the stride. Smaller strides may produce denser latent grids that better preserve object boundaries, but at the cost of increased computation and memory usage.

\section{Conclusion}
The study proposes a method for adding independently learned representations into detection pipelines through backbone augmentation. Our findings show that latent spaces guided by annotations can serve as valuable augmentations in task-driven models.

Future work will extend this framework by developing a continuous latent space method that encodes continuous relationships rather than discrete similarities. Another natural extension would be to develop a latent space method that can model scenes with multiple objects directly by constructing a joint annotation with all object annotations to obtain a single embedding for the scene.


\small
\bibliographystyle{IEEEtranN}
\bibliography{references}

\end{document}